\pdfoutput=1
\documentclass[11pt]{article}

\usepackage[]{EMNLP2023}

\usepackage{times}
\usepackage{latexsym}

\usepackage[T1]{fontenc}
\usepackage[utf8]{inputenc}

\usepackage{microtype}

\usepackage{inconsolata}
\usepackage{graphicx}
\usepackage{tabularx}
\usepackage{fancyvrb} % Add the fancyvrb package
\usepackage{dirtytalk}
\usepackage{booktabs}
\usepackage{hyperref}

\title{Evaluation of Faithfulness Using the Longest Supported Subsequence}

\author{
  \begin{tabular}{ccc}
    \shortstack{Anirudh Mittal\\
    Meta AI\\
    anirudhmittal@meta.com} 
   \shortstack{Timo Schick\\
    Meta AI\\
    schick@meta.com} 
    \shortstack{Mikel Artetxe\\
    Reka AI\\
    mikel@reka.ai} \\
   \shortstack{Jane Dwivedi-Yu\\
    Meta AI\\
    janeyu@meta.com}
\end{tabular}
    }
\author{
Anirudh Mittal \\ Meta AI \\ anirudhmittal@meta.com
\And 
Timo Schick \\ Meta AI \\ schick@meta.com
\And
Mikel Artetxe \\ Reka AI \\ mikel@reka.ai
\And
Jane Dwivedi-Yu \\ Meta AI \\ janeyu@meta.com
}

\begin{document}
\maketitle
\begin{abstract}
As increasingly sophisticated language models emerge, their trustworthiness becomes a pivotal issue, especially in tasks such as summarization and question-answering. Ensuring their responses are contextually grounded and faithful is challenging due to the linguistic diversity and the myriad of possible answers. In this paper, we introduce a novel approach to evaluate faithfulness of machine-generated text by computing the longest noncontinuous substring of the claim that is supported by the context, which we refer to as the Longest Supported Subsequence (LSS). Using a new human-annotated dataset, we finetune a model to generate LSS. We introduce a new method of evaluation and demonstrate that these metrics correlate better with human ratings when LSS is employed, as opposed to when it is not. Our proposed metric demonstrates an 18\% enhancement over the prevailing state-of-the-art metric for faithfulness on our dataset. Our metric consistently outperforms other metrics on a summarization dataset across six different models. Finally, we compare several popular Large Language Models (LLMs) for faithfulness using this metric. We release the human-annotated dataset built for predicting LSS and our fine-tuned model for evaluating faithfulness.\footnote{The data, code and models can be found here: \href{https://github.com/facebookresearch/lss_eval}{facebookresearch.com/lss\_eval} } 
\end{abstract}

\section{Introduction}
\label{sec:introduction}

\begin{figure}[h]
  \centering
\includegraphics[width=\columnwidth]{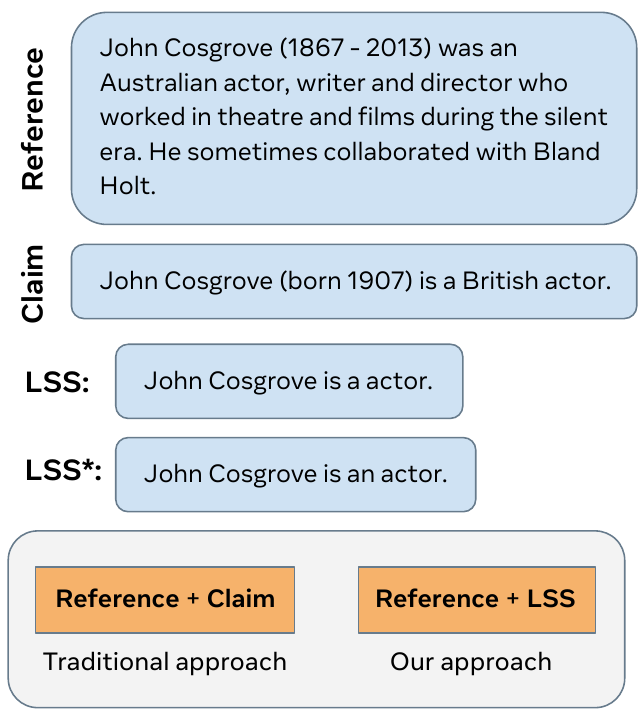}
  \caption{Our proposed new method for evaluating the extent to which the claim is grounded in the reference. Traditional metrics use only the reference and the claim. We generate the Longest Supported Subsequence (LSS) and evaluate it using the claim and the LSS. We also evaluate using LSS*, which is the grammatically correct version of the LSS. We observe that the correlation with human ratings is higher when using LSS.}
  \label{fig:metric}
\end{figure}

The increasing adoption of generative models in mainstream applications has led to their widespread use in tasks such as long-form creative story generation \citep{schick-schutze-2021-shot}, summarization \citep{goyal2023news}, and question answering \citep{pereira2022visconde}. One problem that LLMs suffer from is hallucinations---generating content that is not grounded in the given input or reference \citep{narayan-etal-2018-dont}. This can present problems for tasks like claim verification and summarization, where unsupported information can be detrimental to the application. It is crucial to evaluate the \textit{extent} or degree to which a model deviates or hallucinates from the provided context in order to assess and counteract the issue. In this work, we aim to create a new metric to automatically measure the faithfulness of text generated by models. 

Research on faithfulness can be broadly divided into mitigation and assessment. Methods for reducing hallucinations and improving faithfulness include using faithfulness score to sample generations (\citealp{chen-etal-2021-improving}, \citealp{dong-etal-2020-multi}, \citealp{liu-liu-2021-simcls}, \citealp{ladhak-etal-2022-faithful}) or using the score as model input during training (\citealp{goyal-durrett-2021-annotating}, \citealp{nan-etal-2021-entity}, \citealp{cao-wang-2021-cliff}, \citealp{wan-bansal-2022-factpegasus}, \citealp{zhang-etal-2022-improving-faithfulness}, \citealp{xiao2022entitybased}). In terms of assessment, prior works have studied defining the types of hallucinations, their occurrence and potential sources (\citealp{maynez-etal-2020-faithfulness}, \citealp{pagnoni-etal-2021-understanding}, \citealp{van-der-poel-etal-2022-mutual}). Although various metrics such as n-gram based (ROUGE \citep{lin-2004-rouge}, BLEU \citep{papineni-etal-2002-bleu}), transformer-based (BERTScore \citep{zhang2020bertscore}), and QA-based metrics (QuestEval \citep{scialom-etal-2021-questeval}) have been utilized over the years, they often prove to be brittle and fail to clarify which parts of the generated text are hallucinations. Their poor correlation with human ratings also makes them unsuitable for large-scale automatic evaluation \citep{koto2022ffci}.

We introduce a novel approach to automatically highlight portions of the claim that are supported by a given context. More concretely, given a reference (the contextual paragraph(s)) and a claim (the generated sentence that should be grounded in the reference) as shown in Figure~\ref{fig:metric}, we produce its Longest Supported Subsequence (LSS) using a model trained on a newly collected dataset of reference, claim, and LSS pairs. We subsequently illustrate that employing the computed LSS to assess faithfulness offers a better correlation with human judgments compared to using only the reference and claim. To validate this method, we compare the correlation of our metric with human judgment on summaries generated by various models.  Finally, we compare the faithfulness of multiple advanced LLMs. This comparison is useful for researchers and developers to pick a model for their applications. Our experiments and results can be found in Section \ref{sec:experiments}. We make available both the dataset and the finetuned model for generating LSS in the hopes that this will advance future work in measuring faithfulness of generated text and analyzing when and why hallucinations may occur. 

To summarize our contributions in this work:
\begin{enumerate}
    \item We create and release a new dataset and a lightweight model to predict the longest supported subsequence of a claim given a contextual reference. We also release our LSS* model, where the LSS* is the grammatically-correct version of the LSS. This model can potentially be used to rewrite a claim, containing only portions that are supported by the reference document.
    \item We show that using LSS can lead to significant gains in correlation with human judgment, beating the current state-of-the-art by 18 points on our test data set. We further evaluate our approach on a different task and dataset by comparing generations from 6 different models. 
    \item We compare models like GPT-3.5 \citep{ouyang2022training}, LLaMA \citep{touvron2023llama}, Cohere \citep{cohere} and Flan T5 \citep{chung2022scaling} on 3 different datasets for faithfulness.  We release our metric to support automatic evaluation of text. 
\end{enumerate}

% We posit that our metric's superior correlation with human ratings can play a pivotal role when used for automated model feedback, contributing to the faithfulness of our models. We discuss ideas to further increase this correlation and leverage it to mitigate the challenge of hallucinations. 

\section{Related work}

\paragraph{N-gram Metrics} Due to their simplicity, string overlap-based evaluation metrics have become the standard method for automatically evaluating summarization. ROUGE \citep{lin-2004-rouge} measures the overlap between a generated and reference summary in terms of unigram or bigram overlap (ROUGE-1 and ROUGE-2, respectively), or longest common subsequence (ROUGE-L). BLEU \citep{papineni-etal-2002-bleu} is a precision based metric that has been widely adopted for the task of machine translation and also used as a baseline for evaluation of summarization.

\paragraph{Factuality Checking} Perhaps the most similar to our work is FactCC \citep{kryscinski-etal-2020-evaluating}, which proposed a weakly-supervised, model-based approach for verifying factuality in abstractive summaries.  The training data is automatically generated based on transformation rules that include paraphrasing, entity and number swapping, pronoun swapping, sentence negation, and noise injection.  The goal is to estimate $P(y|A,c)$ where $y$ is a binary label of CORRECT and INCORRECT; $A$ is the reference, and $c$ is the transformed sentence (the claim or summary); and additional span selection heads are utilized for highlighting the supported spans. The training is extended by allowing the model to not only classify the claim consistency but also highlight a span in the source article as the supporting evidence (and denote this model as FactCCX). Unlike FactCC, our dataset is human annotated and more balanced, which leads to better LSS generations. Our approach also centers on a more granular classification, rather than binary label prediction, resulting in increased accuracy as outlined in Section \ref{sec:model}.

\paragraph{Model-based Metrics}

Other approaches incorporate general-purpose string similarity metrics like the unsupervised BERTScore \citep{zhang2020bertscore} and the supervised STS-Score, which receives training from successive SemEval tasks' STS data \citep{agirre-etal-2012-semeval}. \citet{zhang2020bertscore} proposed BERTScore as a means of computing the similarity between BERT token embeddings of system and reference texts. \citet{zhao-etal-2019-moverscore} proposed MoverScore as the Euclidean distance between two contextualized BERT representations. Contextualized word embeddings, such as BERTScore, have been shown to be strong metrics for evaluation in domains like machine translation (\citep{mathur-etal-2019-putting}; \citet{zhang2020bertscore}). Question-answer oriented approaches have been utilized for more than a decade (\citet{wang-etal-2020-asking}, \citet{durmus-etal-2020-feqa}). These works focus on generating questions from summaries and then generating their answer from source documents, typically comparing precision. QuestEval \citep{scialom-etal-2021-questeval} focuses on using a combination of precision and recall metrics to improve the correlation with human judgement. More recently, QAFactEval \ref{fabbri2022qafacteval} combines the best of question answering and entailment metrics. However, question generation and answering is often computationally expensive \citep{koto2022ffci} and critically depends on resources that are often unavailable in languages other than English. Among all other approaches, QuestEval and QAFactEval shows the highest correlation with human judgment as reported by recent works, establishing it as the state-of-the-art for automatic faithfulness evaluation.

Our approach on the other hand focuses on finding the fragments of the sentence that are grounded in context. It offers the advantage of giving a more granular scoring and highlighting the part of sentence that are supported and not supported. 
% According to a survey conducted by, \citep{koto2022ffci} ROUGE is used by more than 95\% of papers. Other metrics such as METEOR, BLEU, BERTScore, and MoverScor ere are rarely used.  Interestingly, 64\% of the surveyed papers employed manual evaluation for their model analysis. Nonetheless, manual evaluation is time-consuming and costly. This emphasizes the need for automatic evaluation metrics that have better correlation with human ratings. 

\section{Approach}
\label{sec:approach}
For several tasks, the expectation from machine-generated text is that it is grounded in the provided context or, in other words, \textit{faithful} to the context. For instance, in the task of summarization, language models may exhibit a propensity to either add non-existent information from the input text, known as extrinsic hallucination, or incorrectly manipulate the input information, termed intrinsic hallucination \citep{maynez-etal-2020-faithfulness}. Similar problems can be noticed in other tasks like question-answering and data-to-text \citep{xiao-wang-2021-hallucination}. Our objective is to quantify faithfulness between a reference document and a claim statement using a normalized metric ranging from 0 to 1, where 0 signifies no support from the reference, and 1 indicates complete support. Here, ``support'' means that the information can be concluded to be true based on the information provided in the input reference. For summarization, the claim would be a summary of the reference document. For our approach, we define the Longest Supported Subsequence (LSS), which is the longest subsequence of words from the claim that is supported by the reference, and it need not be a continuous substring. In Figure~\ref{fig:metric}, we show an example, where the claim contains the date of birth ``(born 1907)'' and nationality ``British,'' both of which are not supported by the reference, and are consequently excluded from the LSS.

As can be seen in this example, the LSS does not have to be grammatically correct. As an alternative to the LSS, we also define the \textit{LSS*} which is a grammatically correct and coherent version of the LSS. In the process of writing LSS*, we make sure not to add any extra information. For the above example, ``a actor'' has been replaced by ``an actor'' in the LSS* in order to make it grammatically correct. More such examples explaining LSS and LSS* annotation can be found in Section \ref{appendix:examples}.

As illustrated in Figure~\ref{fig:metric}, our findings indicate that the use of LSS or LSS* in conjunction with the claim exhibits a higher correlation with human faithfulness ratings, compared to traditional comparisons between the reference and the claim. We compare these two settings using the following metrics: ROUGE \citep{lin-2004-rouge}, BLEU \citep{papineni-etal-2002-bleu}, BERTScore \citep{zhang2020bertscore}, FactCC \citep{wan-bansal-2022-factpegasus}, QuestEval \citep{scialom-etal-2021-questeval}. 

% Intuitively, we felt that LSS can be a useful input for calculating faithfulness. Rather than giving Reference and Claim to traditional metrics, giving Reference, Claim and LSS might make them output a score that has a higher correlation with human metrics. Therefore, what we propose here as our metric is an additional layer of generating LSS on top of the traditional metrics. This can be seen in Figure \ref{fig:metric} For now, we use just Claim and LSS as the input as most traditional metrics use only two text inputs. To create a model that can generate LSS we decide to follow a data-driven approach of finetuning a pre-trained model. According to the survey conducted by \citep{koto2022ffci} most recent work has used the source article as the basis for assessing faithfulness, precision, and recall, rather than reference summaries. Intuitively, this is the best practice in human evaluation, especially for faithfulness, as generated summaries can technically contain details not found in reference summaries but are in the source article.

\section{Dataset}
\label{sec:dataset}
To train our model to automatically annotate LSS given the reference and the claim, we create a high-quality dataset using human annotations. This section describes the process of collecting the dataset and its properties. 

\begin{figure}[h]
  \centering
  \includegraphics[width=0.45\textwidth]{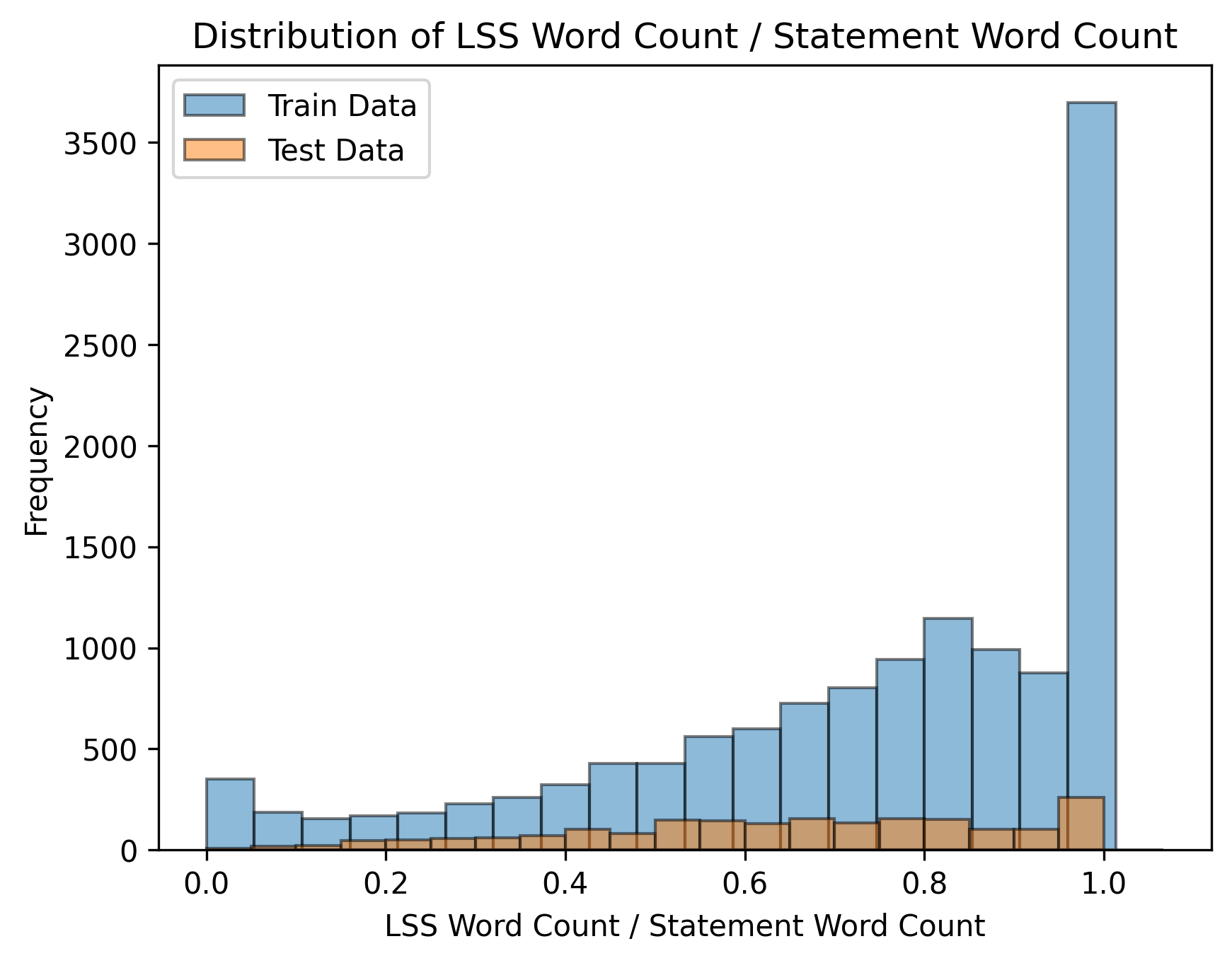}
  \includegraphics[width=0.45\textwidth]{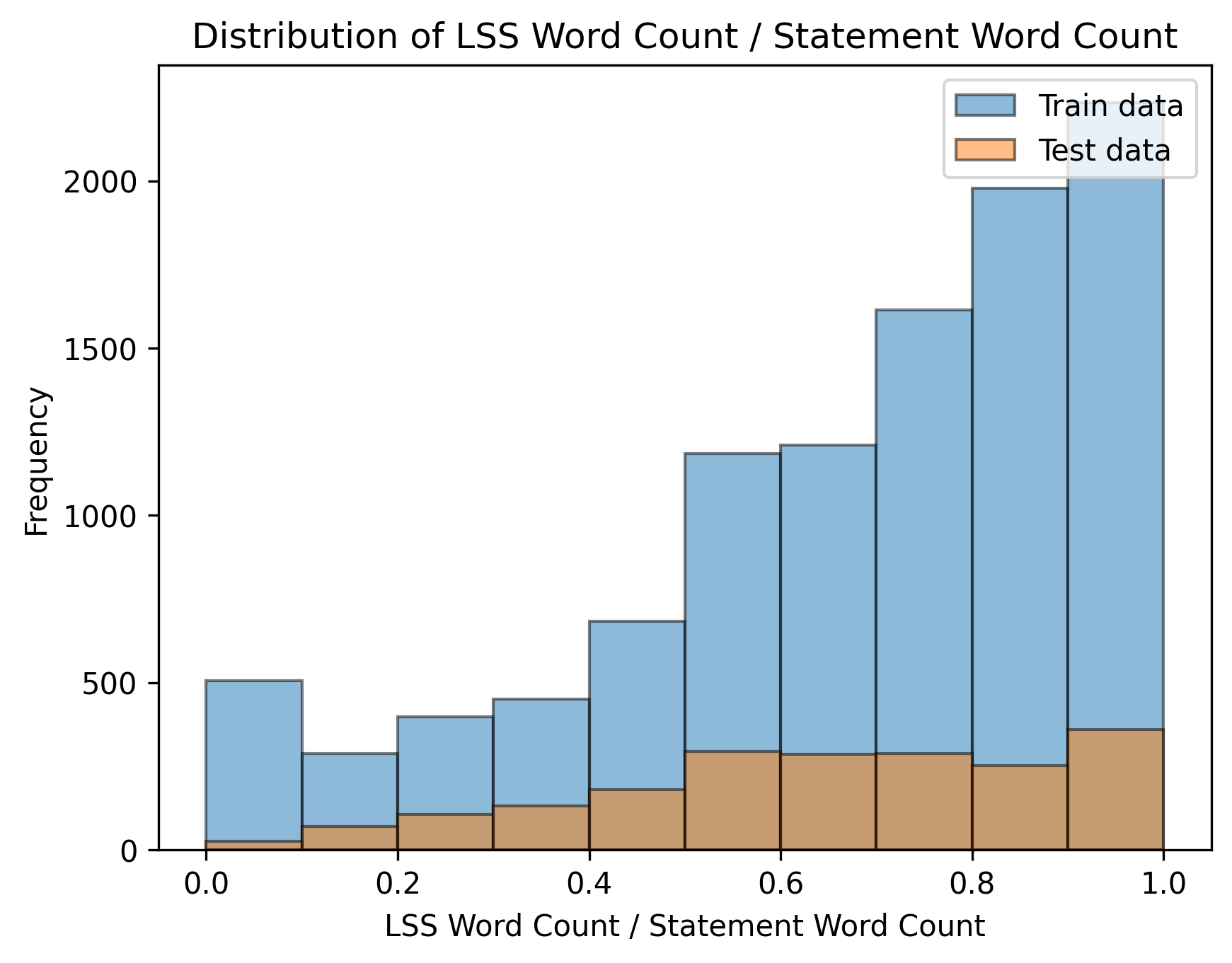}
  \caption{Distribution of word count in LSS divided by word count in the claim for our train and test data. 0 represents an empty LSS (i.e., none of the claim is supported). 1 signifies that the entire claim is supported. The image on the top is the collected LSS dataset and the image on the bottom is the balanced version of the dataset used to train the model.}
  \label{fig:distribution}
\end{figure}

\subsection{Collection}
We collect the \textsc{LSS Dataset} of 15k data points for this project. We start with the reference and its corresponding claim obtained from the WAFER dataset~\citep{petroni2022improving}. The WAFER dataset comprises English sentences or claims from Wikipedia articles, along with the text from their corresponding cited references. Several pre-processing steps were done such as removing reference starting mid-sentences, removing special characters and removing white space. After cleaning the dataset, human annotators were asked to annotate the LSS and LSS* of each reference-claim pair. 

\paragraph{LSS and LSS* Annotation}
Out of 15k examples that were manually annotated, 12k are reserved for training, 1k for validation and 2k for test. For the test data, three annotations were obtained for each data point in order to ensure high quality LSS and LSS* annotations. Given these annotations, we tallied the number of instances where per example all three were the same, two matched, or none were identical. These numbers are reported in Table \ref{tab:iaa}. For cases where two or more of the annotations were the same, the majority vote was taken. In cases where all three of the annotations were different (\textasciitilde25\% cases), one of the three annotation was kept using another round of annotations where an annotator was asked which of the three annotation is correct.  

\paragraph{Faithfulness Rating Annotation} Furthermore, annotators were requested to assign a faithfulness score, ranging from 1 (Not supported) to 5 (Completely supported) for each of the 2k test data points. Interannotator agreement was calculated for the rating of faithfulness using Quadratic Weighted Kappa and found to be 0.63, which indicates significant agreement. More details about the data collection process like the task details and annotation costs can be found in Section \ref{appendix:dataset}.

\subsection{Processing}   
As can be seen in the top graph of Figure \ref{fig:distribution}, the collected dataset was imbalanced containing more cases where length of the LSS to length of the claim ratio is 1. We discovered that balancing the data enhances the model's performance. We eliminated 2k data points where the LSS was identical to the claim. The final training data distribution is represented in the bottom graph of the figure.

% \usepackage{booktabs}
% \usepackage{graphicx}
\begin{table}[]
\centering
\begin{tabular}{@{}ccc@{}}
\toprule
\textbf{Case}          & \textbf{LSS} & \textbf{LSS*} \\ \midrule
\textbf{All same}      & 38.03        & 34.97         \\
\textbf{Two same}      & 34.02        & 33.62         \\ 
\textbf{All different} & 27.96        & 31.41         \\ \bottomrule
\end{tabular}
\caption{Inter annotator agreement of similarity of LSS/LSS* annotations in terms of percentage of the total annotations.}
\label{tab:iaa}
\end{table}

\section{Model}
\label{sec:model}
In this section, we describe the model was trained to generate LSS/LSS* and compare it to relevant baselines. 

\subsection{Model details}
We used the 3 billion-parameter pre-trained language model, t5.1.1.lm100k.xl. This is a variant of the T5 transformer that has been fine-tuned to enhance prompt tuning \citep{lester2021power}. 
We used our training dataset to fine-tune the model using the Huggingface Transformers library \citep{wolf-etal-2020-transformers}. The learning objective, given a reference and claim as input, is to generate the LSS/LSS* as output. More details about the hyperparameters and the prompt used can be found in Section \ref{appendix:hyperparameters}. We also experiment with using several dataset combinations including the FactCC dataset. We found that the model performed the best using the balanced version of our dataset. Details of these experiments are described in Section \ref{appendix:otherdatasets}

\subsection{Model evaluation}
For evaluating the model performance, we used the model to generate the LSS for each reference-claim pair in the test dataset. We compare our fine-tuned model's performance with state-of-the-art models, such as ChatGPT (gpt-3.5-turbo) \citep{ouyang2022training} and Flan T5 XL \citep{chung2022scaling}. We experimented with several prompts to attain optimal performance with both these models. Using the human-written LSS, we assess the output of these three systems using ROUGE, BLEU and word-level metrics, for which we computed the precision, recall, and F1 score for the words in the prediction and the gold answer. Our fine-tuned model's performance surpasses that of the baselines, as seen in Table \ref{tab:model}. The comparatively lower performance of ChatGPT and Flan T5 can often be attributed to their failure to follow instructions in several instances. Specifically, the outputs were longer than the claims due to the introduction of new words not present in the original claim, which occurred despite explicitly forbidding otherwise in the instructions and showing examples in the prompt. The qualitative results generated by our model can be found in Table~\ref{tab:examples}. Comparable outcomes were observed when this experiment was conducted again with LSS*, the details of which can be found in Section \ref{appendix:model_lss_star}.

\begin{table}[]
\begin{tabular}{@{}cccc@{}}
\toprule
\textbf{Metric} & \textbf{LSS Model} & \textbf{ChatGPT} & \textbf{FlanT5} \\ \midrule
\textbf{ROUGE-1}      & 0.85 & 0.55 & 0.59 \\
\textbf{ROUGE-2}      & 0.76 & 0.38 & 0.45 \\
\textbf{ROUGE-L}      & 0.85 & 0.53 & 0.58 \\ \midrule
\textbf{BLEU}           & 0.65 & 0.27 & 0.32 \\ \midrule
\textbf{Precision} & 0.84 & 0.56 & 0.61 \\
\textbf{Recall}    & 0.93 & 0.65 & 0.66 \\
\textbf{F1}        & 0.86 & 0.57 & 0.60 \\ \bottomrule
\end{tabular}
\caption{Evaluation of our LSS model (T5 XL fine-tuned on the \textsc{LSS Dataset}) for the task of generating LSS given the reference and the claim against two state-of-the-art LLMs. }
\label{tab:model}
\end{table}

\subsection{Model generations}

To elucidate this approach, we present an example from each of the five categories of faithfulness defined for our evaluations: ``Not supported'', ``Slightly supported'', ``Partially supported'', ``Almost supported'' and ``Completely supported''. In Table \ref{tab:examples}, we show for a reference-claim pair, the human-written LSS (LSS gold), model-generated LSS (LSS model), and the scores generated using our best performing metric LSS-BLEU and the human rating. The LSS is empty when there is no faithfulness, indicating that the entire claim is not supported by the reference. Similarly, the LSS is equivalent to the claim when there is complete faithfulness, indicating that all of the claim is supported. As observed, the LSS generated by the model does not always align with the LSS crafted by human annotators. The model gets confused with multiple event details that are mentioned, like marriage year in example 3 and time of death in example 4. It manages to get it right when there's a single detail that needs to be confirmed, like date of birth in example 2. It also manages to tackle confusing reference documents like ``third child'' in example 3 and ``villian vs antagonist'' in example 5.

\begin{table*}
\small
% \begin{tabularx}{\textwidth}{@{} *1{X} *5{X} @{}}
\begin{tabularx}{\textwidth}{p{45mm}p{25mm}p{22mm}p{22mm}cp{10mm}}
\toprule
\textbf{Reference} &
  \textbf{Claim} &
  \textbf{LSS gold} &
  \textbf{LSS model} &
  \textbf{LSS score} &
  \textbf{Human} \\ \midrule
Development of an Apple smartphone began in 2004 as the highly confidential ``Project Purple''. Then-Apple CEO Steve Jobs steered the original focus away from a tablet (which was later revisited in the form of the iPad) towards a phone. &
  Apple CEO and Co Founder, Steve Jobs was fired from the company in 1985. &
  - &
  Apple CEO and Co Founder, Steve Jobs &
  0.5 &
  1- Not supported \\ \midrule
Chiang Chung ``CC'' Mei (born 4 April 1935) is Ford Professor of MIT, known for his contribution. He received the Moffatt–Nichol Award in 1992 and the International Coastal Engineering Award in 1995, both from the American Society of Civil Engineers. &
  Chiang C. Mei (; born December 1, 1949) is a Taiwanese academic and politician. &
  Chiang C. Mei is a academic. &
  Chiang C. Mei is a academic. &
  0.3 &
  2 - Slightly Supported \\ \midrule
On May 19, 2012, Kendall got married in the grounds of his mansion. Their second daughter, Iverson, was born in August 2017. On September 21, 2022, Kendall announced that he and Rava are expecting their third daughter, due in 2023. &
  They got married in July 2015 and got separated 4 years later after having 2 kids. &
  They got married having 2 kids &
  They got married in July 2015 and having 2 kids. &
  0.65 &
  3 - Partially Supported \\ \midrule
On 8 September 2022, Buckingham Palace released a statement. She died at 15:10 BST, with her death announced to the public at 18:30, setting in motion Operation London Bridge and, because she died in Scotland, Operation Unicorn. &
  The Queen died on the 8th of September at 18:30, setting Operation London Bridge in motion. &
  The Queen died on the 8th of September, setting Operation London Bridge in motion. &
  The Queen died on the 8th of September at 18:30, setting Operation London Bridge in motion. &
  1 &
  4- Almost supported \\ \midrule
The Charan Raj Returns! with Paisa Charan Raj will be seen in a role of a politician in the movie. He is doing the role of the main villain after a long gap. &
  Charan Raj was selected to play the main antagonist of the film. &
  Charan Raj was selected to play the main antagonist of the film. &
  Charan Raj was selected to play the main antagonist of the film. &
  1 &
  5 - Completely supported \\ \bottomrule
\end{tabularx}
\caption{Examples of reference and claim used from the dataset along with the LSS written by human annotators (LSS gold), LSS generated by our model (LSS Model), score generated by our metric and the human rating. Our metric here refers to LSS-BLEU, it ranges from 0 to 1. Human rating is an integer label between 1 to 5 where 5 means completely faithful.}
\label{tab:examples}
\end{table*}

\section{Evaluation and results}
\label{sec:experiments}

% \usepackage{booktabs}
% \usepackage{graphicx}
\begin{table*}[t]
\centering
\begin{tabular}{@{}c|c|cc|cc@{}}
\toprule
\multicolumn{1}{c}{} &
  \multicolumn{1}{c}{\textbf{Reference-Claim}} &
  \multicolumn{2}{c}{\textbf{LSS-Claim}} &
  \multicolumn{2}{c}{\textbf{LSS*-Claim}} \\ \midrule
\multicolumn{1}{c}{\textbf{Metric}} &
  \multicolumn{1}{c}{---} &
  \multicolumn{1}{c}{\textbf{Human}} &
  \multicolumn{1}{c}{\textbf{Generated}} &
  \multicolumn{1}{c}{\textbf{Human}} &
  \multicolumn{1}{c}{\textbf{Generated}} \\ \midrule
\textbf{ROUGE}                  & -0.03        & \multicolumn{1}{c}{\textbf{0.86}} & 0.46          & \multicolumn{1}{c}{\textbf{0.86}} & 0.47          \\ 
\textbf{BLEU}                   & -0.07        & \multicolumn{1}{c}{0.85}          & \textbf{0.48} & \multicolumn{1}{c}{0.84}          & \textbf{0.49} \\ 
\textbf{BERTScore} & 0.19         & \multicolumn{1}{c}{0.28}          & 0.27          & \multicolumn{1}{c}{0.28}          & 0.20          \\ 
\textbf{FactCC}                 & 0.05         & \multicolumn{1}{c}{0.68}          & 0.24          & \multicolumn{1}{c}{0.66}          & 0.26          \\ 
\textbf{QuestEval}              & \textbf{0.3} & \multicolumn{1}{c}{0.54}          & 0.33          & \multicolumn{1}{c}{0.55}          & 0.30          \\ 
\textbf{QAFactEval}              & 0.11 & \multicolumn{1}{c}{0.45}          & 0.36          & \multicolumn{1}{c}{0.45}          & 0.35         \\ \bottomrule
\end{tabular}

\caption{Results on the test split of \textsc{LSS Dataset}. Each column is the metric calculation between the reference and the claim. \textit{Human} is between the claim and the human-written LSS while \textit{Generated} is between the claim and the model-generated LSS. The highest value for each column is in bold.}
\label{tab:metric}
\end{table*}

We first demonstrate our approach for measuring the factuality of a claim for a given reference on the \textsc{LSS Dataset}. Second, we evaluate our approach on the out-of-domain task of summarization (summaries instead of claims) using summaries generated by several models and the XSum dataset~\citep{narayan-etal-2018-dont}. Lastly, we apply our metric to evaluate the faithfulness of various LLMs.
%  \citet{koto2022ffci} found that absolute scoring is more common than relative evaluation.  Absolute https://www.overleaf.com/project/63b7118cc74f26403f01786fbenchmark is conducted by asking annotators to evaluate system-generated summaries based on a numeric scale,  in isolation from any other summaries.  With relative evaluation, on the other hand, annotators are asked to directly rank summaries generated by different methods. We follow the absolute scoring approach for this work.

\subsection{\textsc{LSS Dataset}}
We applied and compared our approach on the test split of the \textsc{LSS Dataset}, which consists of 2,000 samples. 
% A survey by \citep{koto2022ffci} reveals that most recent studies utilize the source article or reference, as the assessment basis for faithfulness, instead of the reference summaries. 
Our evaluation draws comparisons under two settings: (1) using the reference and the claim, (2) using the generated LSS and the claim, and (3) human-written LSS and the claim. 
% , establishing an upper bound for the correlation with human ratings of faithfulness in our methodology. 
To evaluate setting (2), for each reference-claim pair in our dataset, we used our fine-tuned models to generate the corresponding LSS/LSS*. Setting (3) presumably represents the optimal LSS that could be composed for a given reference-claim pair, and helps account for generation errors in setting (2). We use the metrics BLEU, ROUGE, BERTScore, FactCC, QuestEval and QAFactEval.  For the three different cases--- Traditional (reference-claim), Model (claim-generated LSS), and Human (claim-human LSS)---we calculate Pearson and Spearman correlation coefficients with human ratings for faithfulness, where a higher correlation value is desirable.
% As human ratings are regarded as the gold standard, a higher correlation value is desirable. 
These results are reported in Table \ref{tab:metric}. 

As presented in Table \ref{tab:metric}, using the traditional evaluation of reference-claim, the maximum correlation attainable with human ratings on the \textsc{LSS Dataset} is 0.3 using the QuestEval metric. Despite their widespread use in faithfulness evaluation, N-gram based metrics, such as BLEU and ROUGE, demonstrate negative correlation with human judgment. 
These results are in line with previous works on faithfulness (\citealp[]{koto2022ffci, scialom-etal-2021-questeval}). Using our approach, the correlation reaches 0.48 and 0.49 using LSS and LSS*, respectively. Correlation with human judgement using LSS is significantly higher than using the reference instead. Values in the \textit{Human} columns represent the correlation with human judgement using the human-annotated LSS, achieving values as high as 0.86. The gap between our value of 0.48 and 0.86 can be attributed to noise in the LSS generation and demonstrates that further improvements in this space can be made. Metrics like QuestEval and QAFactEval are designed for question generation and answering based on the reference document. When used for claim-LSS comparison they might not give substantial benefits as the claim is often a single sentence.

\subsection{Summarization} 
\label{sec:ood}

\begin{figure*}[]
  \centering
\includegraphics[width=\textwidth]{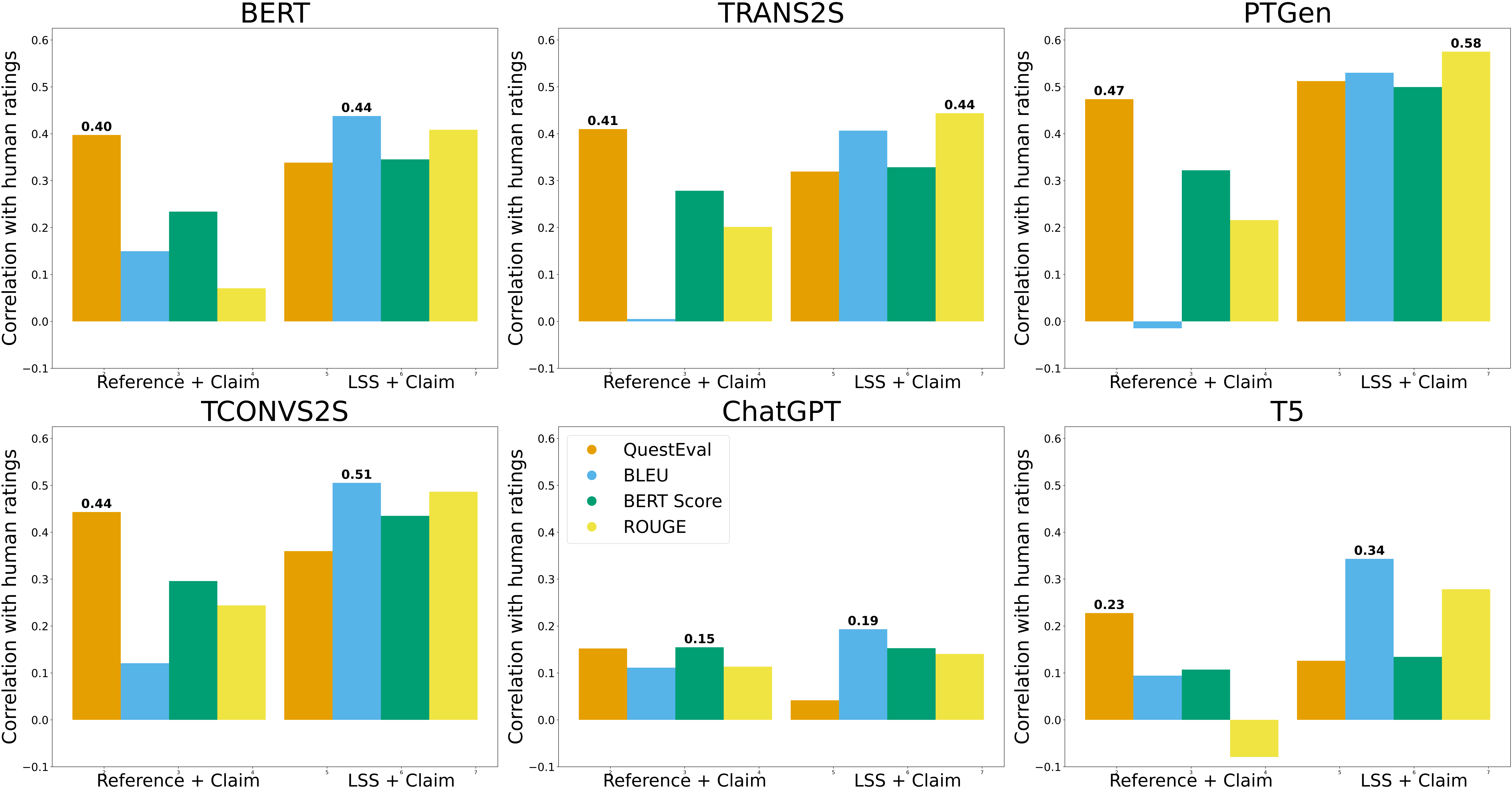}
  \caption{Performance of our metric for the task of summarization. We evaluate summaries written by six different models. \textit{Reference + Claim} here refers to performing evaluation using the reference and the claim. \textit{LSS + Claim} refers to using the claim and the LSS. The highest value for the metrics computed using the reference-summary and the LSS-summary pairs have also been labelled.}
  \label{fig:ood}
\end{figure*}

Given that our LSS-generating model was trained on the same domain as the LSS test dataset, we test our approach on a different domain and dataset, focusing on the task of summarization. For this, we used the XSum dataset \citep{narayan-etal-2018-dont} which is one of the most widely used datasets for summarization evaluation. This dataset encompasses BBC articles, each accompanied by a professionally written one-line summaries. We employ the 2,000 annotations released by \citet{maynez-etal-2020-faithfulness}, which contain human ratings of factuality for summaries generated by four neural models: pointer generator network (“PG”: \citet{see-etal-2017-get}), Topic-aware convolutional sequence-to-sequence model (“TCONV”: \citet{narayan-etal-2018-dont}), a transformer-based model (“TRANS2S”: \citet{vaswani2017attention}),  and BERT (\citet{devlin-etal-2019-bert}, \citet{liu2019text}.  In this case, the documents serve as the reference, and the summaries act as the claim. Considering the dataset's release in 2018 and its reliance on relatively dated models, we sought to evaluate the efficacy of our approach with more recent LLMs. We generate summaries by ChatGPT \citep{ouyang2022training} and T5 XXL \citep{2020t5} and compare these results. For 500 generations from each model, we obtain human annotations for faithfulness on a scale from 1 (Not supported) to 5 (Completely supported), just as was done for \textsc{LSS Dataset}. 

To compare our measurement of faithfulness, we use our model to generate LSS for each pair of documents and model-generated summaries. As our model is trained on a maximum input length of 512 tokens, we remove inputs that exceed this limit (\textasciitilde12\%). We then compare the human correlation of metrics computed using the reference and the claim (Reference-Claim) versus using the claim and the LSS (LSS-Claim). The results of this comparison are depicted in Figure \ref{fig:ood}.

Our approach works well for the task of summarization, as illustrated in Figure~\ref{fig:ood}. For all the 6 models, the highest correlation obtained using LSS-claim outperforms that which uses reference-claim. When using reference-claim, QuestEval exhibits superior performance in all six instances. When using LSS-claim, BLEU performs the best in 4 out of 6 cases, closely followed by ROUGE. The performance of these four metrics on this dataset aligns closely with their performance on the \textsc{LSS Dataset}, suggesting this approach can be applied to out-of-domain datasets.

\subsection{Comparing Models}

In this experiment, we evaluate generations of state of the art LLMs using our metric. We choose 5 popular and useful LLMs  - ChatGPT, Cohere, LLaMa, Flan T5, Vicuna to generate summaries for 1,000 documents each from XSum and CNN-DM. Additionally, we also test them on our LSS test dataset. Using our model, we calculate the LSS from the document and the generated summaries. Subsequently, we compute the faithfulness score by applying our best-performing metric, LSS-BLEU. These results are reported in Table \ref{tab:llms}.

When comparing the faithfulness of LLMs, ChatGPT demonstrates superior performance with fewer hallucinations observed in its summaries. Cohere and Flan T5 closely trail in performance. Smaller models like LLaMa and Vicuna appear to produce summaries that are the least faithful. An empirical analysis indicates that this is usually due to exhibiting incoherence and producing gibberish at times. %This shows that instruction-tuned model are more faithful for summarization and generate grounded summaries. 

\begin{table}[]
\centering
\begin{tabular}{@{}cccc@{}}
\toprule
\textbf{Model} & \textbf{Xsum} & \textbf{CNN-DM} & \textbf{LSS} \\ \midrule
\textbf{Cohere}     & 0.83            & 0.92              & 0.90               \\
\textbf{ChatGPT}      & \textbf{0.98}            & \textbf{0.96}              & \textbf{0.98}               \\
\textbf{Flan T5 XL}    & 0.73            & 0.90             & 0.88               \\
\textbf{LLaMa 7B}      & 0.16            & 0.17              & 0.30              \\
\textbf{Vicuna 13B}        & 0.07            & 0.04              & 0.16               \\ \hline
\end{tabular}
\caption{Comparison of faithfulness of LLMs using our evaluation. The `claim' in this setting is the model-generated summaries using documents from XSum, CNN-Daily Mail and the \textsc{LSS Dataset}.}
\label{tab:llms}
\end{table}

\section{Conclusion}
\label{sec:conclusion}
This paper proposes a new metric demonstrating state-of-the-art correlation with human judgment for faithfulness. Our approach builds on top of existing metrics using the concept of the longest supported subsequence. We conduct several experiments to verify the applicability of the approach. We demonstrate the use of this metric by comparing several state-of-the-art LLMs for the task of faithfulness. As part of this work, we release our dataset which can be useful for the community for further research. We release our model, which will allow researchers to use our metric to compare the faithfulness of their proposed model. 

A notable application of our metric lies in providing automated feedback for model generations, for example, in the context of techniques like RLHF and PPO to reduce hallucination, which we leave for future work. 
% We believe a model with higher human correlation like ours, can make this process considerably more efficient. We leave this as future work. 
Another area for improvement is the generation of LSS. As shown in Table \ref{tab:metric}, the model performance is better than the baselines, but there is still a large gap in the values obtained using the generated versus human-written LSS, stemming from the model's ability to completely accurately generate the LSS. Improving the ability to generate a better LSS would lead to a considerable improvement in correlations with human judgement.

\section*{Limitations and Ethical Concerns}

Because we use a dataset that is taken from Wikipedia content, there may be biases inherited by our LSS-generating model. Additionally, given that our approach builds upon existing metrics, it inevitably also inherits their potential flaws or limitations. For instance, current metrics struggle to accurately interpret negation cases. In such scenarios, our model has exhibited similar performance. Compared to the traditional evaluation setting of using reference-claim, our approach may require more time for evaluation due to the generation of LSS. Given that our training utilized a Wikipedia-based dataset, the models perform well on texts with a similar structure and vocabulary, but may not perform as effectively on substantially different text structures, such as those found on Reddit and other social media platforms.

% \section*{Ethics Statement}
% Scientific work published at ACL 2023 must comply with the ACL Ethics Policy.\footnote{\url{https://www.aclweb.org/portal/content/acl-code-ethics}} We encourage all authors to include an explicit ethics statement on the broader impact of the work, or other ethical considerations after the conclusion but before the references. The ethics statement will not count toward the page limit (8 pages for long, 4 pages for short papers).

% We use a dataset that is taken from Wikipedia content. This implies that our modal may inherit biases that already exist in the dataset.

\bibliography{anthology,custom}
\bibliographystyle{acl_natbib}

\appendix

\section{Appendix}
\label{sec:appendix}

\subsection{LSS examples}
\label{appendix:examples}

In this section, we explain the process of writing LSS for the examples mentioned in Table \ref{tab:examples}. 
\begin{enumerate}
    \item \textbf{Example 1:} The reference does not contain any information regarding the firing of Steve Jobs, instead is about a product launch. Therefore, none of the claim is supported.
    \item \textbf{Example 2:} Chiang's date of birth is incorrect in the claim. There is no mention of politics. There is enough evidence (Professor, Engineering award) to support the ``academic'' claim.
    \item \textbf{Example 3:} The year of marriage is 2012 and not 2015. The separation claim is not mentioned in the reference. The reference mentions the third children as expecting therefore they currently have two children.
    \item \textbf{Example 4:} Everything is supported except the time of death which is 15:10 BST and not 18:30.
    \item \textbf{Example 5:} Everything is supported.
\end{enumerate}

\subsection{Dataset creation}
\label{appendix:dataset}
For the task of dataset creation, one annotator was used for train and validation set. Three annotators were used for test and Experiment \ref{sec:ood}. Annotators were paid 2 USD for each annotation. <Annotators profile?>

Following questions were asked during the annotations:

\begin{enumerate}
    \item \textbf{LSS:} Remove words from the statement until it is fully supported by the text (if none of the statement is supported, then leave this empty)
    \item \textbf{LSS*:} Fix any spelling or grammar errors in the LSS
    \item \textbf{Rating:} On a scale of 1-5, how well is the Claim supported by the Reference?
\end{enumerate}

Pilot test was conducted to align the annotators for writing LSS and LSS*. For the rating task, annotators were shown examples for each of the 5 categories. One time instructions were:

Longest supported subsequence (LSS): In this task, you will be given a statement and a reference. You have to remove words from the LSS that are not supported by the reference. Here ``supported'' means if it can be inferred after reading the reference. The LSS does not need to be a continuous copy of the statement! It may have portions in the middle that are deleted. In writing the LSS, disregard evaluating whether what is stated in the reference is true or not. If not empty, you should try to not remove the subject in the statement (See the 4th example).

LSS*: After obtaining LSS, you can add filler words to make the sentence grammatically correct or coherent. This rewritten version is called LSS*. It may or may not be the same as LSS. Please refer to the examples below to see how to write LSS and LSS*.

\subsection{Hyperparameters}
\label{appendix:hyperparameters}
We iterate over model hyperparameters to find the optimal configuration. Final parameters are:
\begin{enumerate}
    \item Epochs: 10
    \item Number of beams: 5
    \item Input max token length: 528
    \item Output max token length: 128
\end{enumerate}

Prompt for input for training the model was: \begin{verbatim}
    'Reference: <reference>\n Claim: <claim>\n Output:'
\end{verbatim}

\subsection{LSS* model evaluation}
\label{appendix:model_lss_star}

We compare the performance of our LSS* generating model against ChatGPT and Flan T5 similar to Table \ref{tab:model}. Just as for LSS, our LSS* model outperforms both the baselines.

\begin{table}[]
\begin{tabular}{@{}cccc@{}}
\toprule
\textbf{Metric} & \textbf{Our Model} & \textbf{ChatGPT} & \textbf{FlanT5} \\ \midrule
\textbf{ROUGE-1}      & 0.85 & 0.55 & 0.59 \\
\textbf{ROUGE-2}      & 0.76 & 0.38 & 0.45 \\
\textbf{ROUGE-L}      & 0.85 & 0.53 & 0.58 \\ \hline
\textbf{BLEU}           & 0.65 & 0.27 & 0.32 \\ \hline
\textbf{Precision} & 0.84 & 0.56 & 0.61 \\
\textbf{Recall}    & 0.93 & 0.65 & 0.66 \\
\textbf{F1}        & 0.86 & 0.57 & 0.60 \\ \bottomrule
\end{tabular}
\caption{Evaluation of our model for the task of generating LSS* given the reference and the claim against two state-of-the-art LLMs. }
\label{tab:model_lss_star}
\end{table}

\subsection{Other datasets}
\label{appendix:otherdatasets}
Other datasets that were considered for training includes the super version of the \textsc{LSS Dataset}. This dataset contains repetitions to balance the LSS/Claim length distribution. 
\begin{figure}[h]
  \centering
  \includegraphics[width=0.50\textwidth]{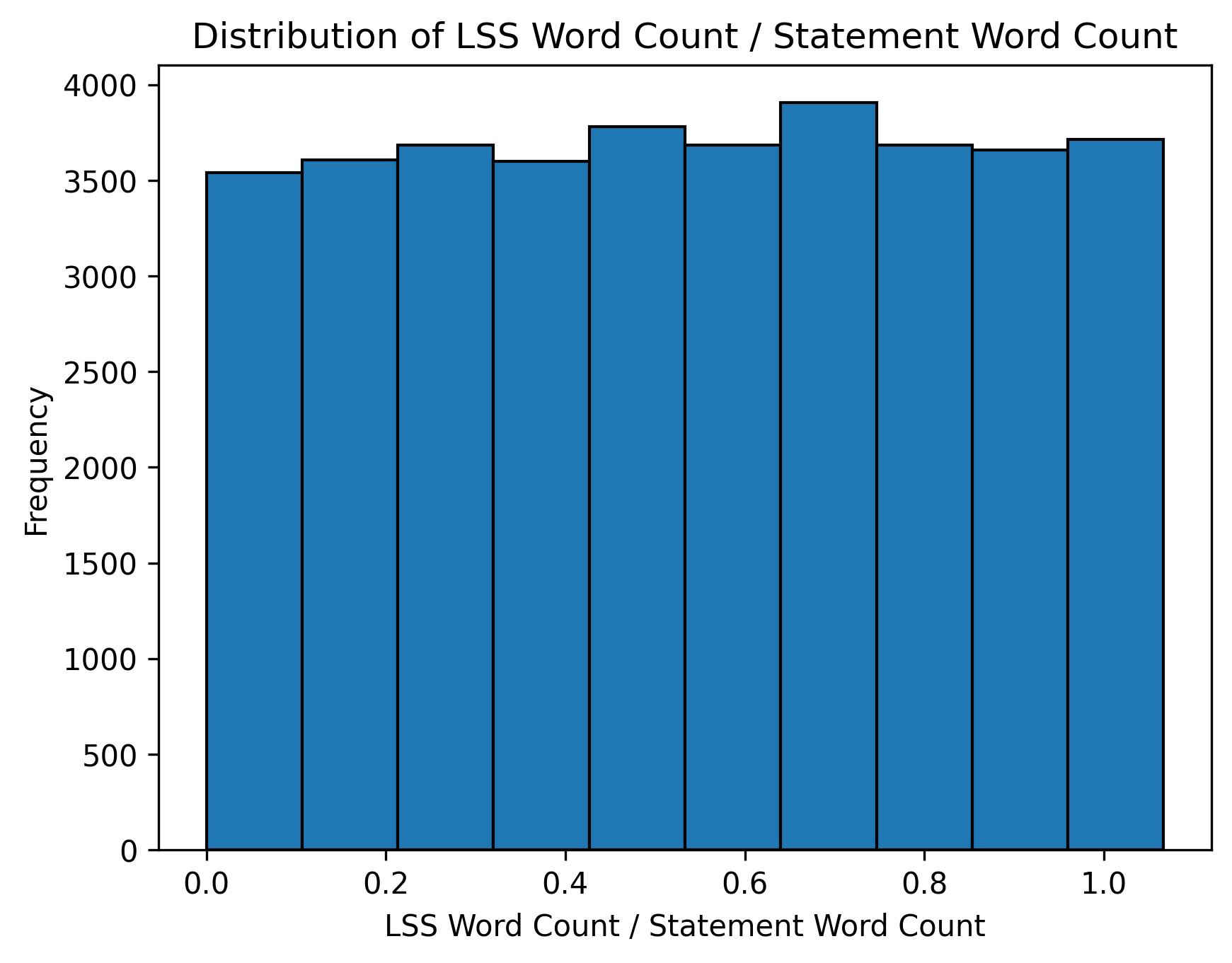}
  \caption{Super version that includes repetitions of datapoints to make it balanced - Distribution of word count in LSS by word count in the claim for our train and test data.}
  \label{fig:super}
\end{figure}

\begin{figure}[h]
  \centering
  \includegraphics[width=0.50\textwidth]{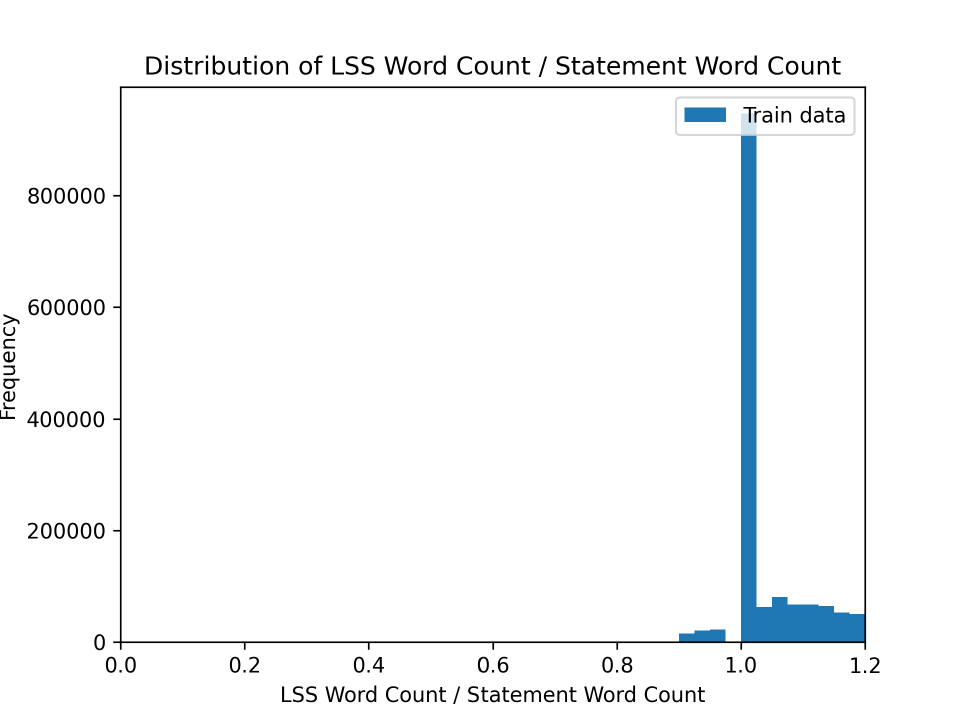}
  \caption{FactCC data. Distribution of word count in LSS by word count in the claim for our train and test data. }
  \label{fig:factcc}
\end{figure}

The distribution of the FactCC data can be found in Fig \ref{fig:factcc}. As can be observed, it is highly inclined towards having LSS same as claim or quite similar to claim as the process involves making changes to a single word or a phrase. 

We consider using the FactCC dataset for training our LSS generation model. In Table \ref{tab:factccdata} we show comparison of model performance using the LSS data, FactCC data and LSS + FactCC data together. We observe that the model trained on LSS performs much better than the one trained on just FactCC. It is important to note that FactCC is 100x times larger than the \textsc{LSS Dataset}. The model trained on both LSS and FactCC performs marginally better than just LSS. Presumably this improvement in accuracy is insignificant when the additonal training is taken into account.

\begin{table}[]
\resizebox{\columnwidth}{!}{%
\begin{tabular}{@{}cccc@{}}
\toprule
\textbf{Metric}         & \textbf{LSS} & \textbf{LSS + FactCC} & \textbf{FactCC} \\ \midrule
\textbf{ROUGE-1 F}      & 0.84         & 0.85                  & 0.54            \\ 
\textbf{ROUGE-2 F}      & 0.75         & 0.77                  & 0.35            \\ 
\textbf{ROUGE-L F}      & 0.84         & 0.85                  & 0.54            \\ \midrule
\textbf{BLEU}           & 0.65         & 0.66                  & 0.25            \\ \midrule
\textbf{Word Precision} & 0.83         & 0.84                  & 0.50            \\ 
\textbf{Word Recall}    & 0.94         & 0.94                  & 0.67            \\ 
\textbf{Word F1}        & 0.86         & 0.87                  & 0.55            \\ \bottomrule
\end{tabular}%
}
\caption{Comparison of model performance when trained on different dataset combinations}
\label{tab:factccdata}
\end{table}

\subsection{Prompt examples}
\label{appendix:prompt}

Several prompts were tried to obtain optimal performance for ChatGPT and FlanT5. The best performing prompts are the following. They include the instructions and a few examples. 

\textbf{LSS:} - 
Instructions: "You are an editor who carefully reads the given text and follows the instructions. In this task, you will be given a Claim and a Reference. You have to remove words from the Claim that are not supported by the Reference. Here "supported" means if it can be inferred after reading the Reference. You can only delete words from the Claim and not allowed to add words. Your generation should never be longer than the Claim."

Examples: "Reference: On 8 September 2022, Buckingham Palace released a statement which read: "Following further evaluation this morning, the Queen's doctors are concerned for Her Majesty's health and have recommended she remain under medical supervision. The Queen remains comfortable and at Balmoral." Elizabeth's four children, her daughters-in-law Camilla and Sophie, and her grandsons William and Harry travelled to Balmoral. She died at 15:10 BST, with her death announced to the public at 18:30, setting in motion Operation London Bridge and, because she died in Scotland, Operation Unicorn. The cause of death was recorded on the death certificate as "old age"
Claim: The Queen died on the 8th of September at 18:30, setting Operation London Bridge in motion.
Output: The Queen died on the 8th of September, setting Operation London Bridge in motion."

"Reference: Chiang Chung "CC" Mei (born 4 April 1935) is Ford Professor of Engineering, Emeritus, at the Department of Civil and Environmental Engineering of Massachusetts Institute of Technology, known for his contributions in fluid mechanics with applications to civil, environmental, and coastal engineering. He has been an Associate Editor of the Journal of Fluid Mechanics. He received the Moffatt–Nichol Award in 1992 and the International Coastal Engineering Award in 1995, both from the American Society of Civil Engineers.
Claim: Chiang C. Mei (; born December 1, 1949) is a Taiwanese academic and politician.
Output: Chiang C. Mei is a academic."

"Reference: Kendall met his future wife, fellow Harvard student Rava, at a frat party during his sophomore year there. They began dating in 2003. In September 2010, Rava, moved into Kendall's rented house in New York. On May 19, 2012, they married in the grounds of his mansion in an event that also celebrated her graduation from medical school. On July 31, 2015, Kendall revealed that they were expecting a baby girl. Their daughter, Sophie Roy, was born on December 1, 2015. Their second daughter, Iverson, was born in August 2017. On September 21, 2022, Kendall announced that he and Rava are expecting their third daughter, due in 2023.
Claim: They got married in July 2015 and got separated 4 years later after having 2 kids.
Output: They got married having 2 kids."
"Reference: "One" is the thirty-first series finale of the British medical drama television series "Casualty", and the 1049th episode of the overall series. The episode is written by Paul Unwin, the show's co-creator, and directed by Jon Sen.
Claim: "One" is the first episode of the fourth series of the British medical drama "Casualty".
Output: "One" is episode of the British medical drama "Casualty"."

\textbf{LSS*} - 
Instructions: "You are an editor who carefully reads the given text and follows the instructions. In this task, you will be given a Claim and a Reference. You have to remove words from the Claim that are not supported by the Reference. Here "supported" means if it can be inferred after reading the Reference. After removing the words, make the sentence grammatically correct and coherent. Only add filler words that do not add a lot of information. "

Examples: 
"Reference: On 8 September 2022, Buckingham Palace released a statement which read: "Following further evaluation this morning, the Queen's doctors are concerned for Her Majesty's health and have recommended she remain under medical supervision. The Queen remains comfortable and at Balmoral." Elizabeth's four children, her daughters-in-law Camilla and Sophie, and her grandsons William and Harry travelled to Balmoral. She died at 15:10 BST, with her death announced to the public at 18:30, setting in motion Operation London Bridge and, because she died in Scotland, Operation Unicorn. The cause of death was recorded on the death certificate as "old age"
Claim: The Queen died on the 8th of September at 18:30, setting Operation London Bridge in motion.
Output: The Queen died on the 8th of September, setting Operation London Bridge in motion."

"Reference: Chiang Chung "CC" Mei (born 4 April 1935) is Ford Professor of Engineering, Emeritus, at the Department of Civil and Environmental Engineering of Massachusetts Institute of Technology, known for his contributions in fluid mechanics with applications to civil, environmental, and coastal engineering. He has been an Associate Editor of the Journal of Fluid Mechanics. He received the Moffatt–Nichol Award in 1992 and the International Coastal Engineering Award in 1995, both from the American Society of Civil Engineers.
Claim: Chiang C. Mei (; born December 1, 1949) is a Taiwanese academic and politician.
Output:Chiang C. Mei is an academic."

"Reference: Kendall met his future wife, fellow Harvard student Rava, at a frat party during his sophomore year there. They began dating in 2003. In September 2010, Rava, moved into Kendall's rented house in New York. On May 19, 2012, they married in the grounds of his mansion in an event that also celebrated her graduation from medical school. On July 31, 2015, Kendall revealed that they were expecting a baby girl. Their daughter, Sophie Roy, was born on December 1, 2015. Their second daughter, Iverson, was born in August 2017. On September 21, 2022, Kendall announced that he and Rava are expecting their third daughter, due in 2023.
Claim: They got married in July 2015 and got separated 4 years later after having 2 kids.
Output:They got married and have 2 kids."
"Reference: "One" is the thirty-first series finale of the British medical drama television series "Casualty", and the 1049th episode of the overall series. The episode is written by Paul Unwin, the show's co-creator, and directed by Jon Sen.
Claim: "One" is the first episode of the fourth series of the British medical drama "Casualty".
Output: "One" is an episode of the British medical drama "Casualty"."

\end{document}